# Deep Learning-Enhanced for Amine Emission Monitoring and Performance Analysis in Industrial Carbon Capture Plants

Lokendra Poudel*[1], David Tincher[1], Duy-Nhat Phan[1], Rahul Bhowmik*[1]

[1]Polaron Analytics, 9059 Springboro Pike, Miamisburg, OH, 45342, USA

*Correspondence to:

Email address: lokendra@polaronanalytics.com
              rahulbhowmik@polaronanalytics.com

**Abstract**

We present data driven deep learning models for forecasting and monitoring amine emissions and key performance parameters in amine-based post-combustion carbon capture systems. Using operational data from the CESAR1 solvent campaign at Technology Center Mongstad, four DL architectures such as Basic Long Short-Term Memory (LSTM), Stacked LSTM, Bi-directional LSTM, and Convolutional LSTM were developed to capture time-dependent process behavior. For emission prediction, models were designed for 2-amino-2-methyl-1-propanol (AMP) and Piperazine emissions measured via FTIR and IMR-MS methods. System performance models target four critical parameters: $CO_2$ product flow, absorber outlet temperature, depleted flue gas outlet temperature, and RFCC stripper bottom temperature. These models achieved high predictive accuracy exceeding 99% and effectively tracked both steady trends and abrupt fluctuations. Additionally, we conducted causal impact analysis to evaluate how operational variables influence emissions and system performance. Eight input variables were systematically perturbed within ±20% of nominal values to simulate deviations and assess their impact. This analysis revealed that adjusting specific operational parameters, such as lean solvent temperature and water wash conditions, can significantly reduce amine emissions and enhance system performance. This study highlights ML not only as a predictive tool but also as a decision-support system for optimizing carbon capture operations under steady-state and dynamic conditions. By enabling real-time monitoring, scenario testing, and operational optimization, the developed ML framework offers a practical pathway for mitigating environmental impacts. This work represents a step toward intelligent, data-driven control strategies that enhance the efficiency, stability, and sustainability of carbon capture and storage technologies.

**Key Word:** Deep Learning, Long-Short Term Memory (LSTM), Amine Emissions, Causal Impact Analysis, Carbon Capture Process

## 1. Introduction

The critical issue of global warming, largely driven by the significant release of carbon dioxide ($CO_2$) from power generation and industrial activities, demands immediate and efficient measures for mitigation (Fawzy et al., 2020; Nunes, 2023). To achieve net-zero carbon emissions by 2050, advancements in technology are essential, particularly in the capture, conversion, transport, storage, and utilization of carbon dioxide (Fam & Fam, 2024; Lau & Tsai, 2023). One key approach is the use of advanced carbon capture plants (CCPs), which work by isolating and storing carbon dioxide to stop it from entering the atmosphere. This method could significantly help lower greenhouse gas emissions and combat climate change, especially in areas like power generation and heavy industry where cutting emissions is particularly difficult. It involves the process of capturing $CO_2$ emissions produced from various power plants and industrial activities before they are released into the atmosphere (Hanson et al., 2025). A common method is amine-based carbon capture, which is a chemical absorption process where $CO_2$ is captured using amines solvents, such as MEA (monoethanolamine), CESAR1 (2-amino-2-methyl-1-propanol and piperazine) (Hume et al., 2022; Tatarczuk et al., 2024). In this process, the flue gas (FG) emitted from industrial sources such as power plants or cement factories is exposed to an amine solvent. The $CO_2$ in the FG reacts with the amine solution, which creates a chemical bond that effectively captures and removes the $CO_2$ from the gas stream (Tatarczuk et al., 2024). The $CO_2$-rich amine solvent is then separated from the gas stream and heated to release the captured $CO_2$, which can be compressed and transported for storage or utilization (Panja et al., 2022). Although solvent-based carbon capture holds promise for enhancing air quality by removing amine gases like $SO_2$, $CO_2$, $CO$, $NH_3$, $NO_2$ etc., it also poses challenges, including solvent degradation and the generation of unwanted by-products of amine compounds (Vevelstad et al., 2022).

Amine-based carbon capture processes present significant environmental concerns by solvent degradation. The degradation of solvent releasing volatile amine compounds into the air, particularly during intermittent operation of power plants (Jablonka et al., 2023). These amine compounds can undergo atmospheric reactions by producing harmful secondary pollutants such as nitrosamines, ozone, and fine particulate matter ($PM_{2.5}$), posing serious risks to air quality and human health (Rochelle, 2024). The potential for such emissions underscores the need for a comprehensive understanding of chemical reactions, reaction kinetics, and the interactions between process parameters in plant design, control, and optimization. Traditional process models often assume steady-state operation, which limits their applicability in real-world scenarios where plants must adapt to variable conditions, especially with the increasing integration of intermittent energy sources. There is a critical need for advanced methods that can capture the dynamic, nonlinear, and multivariate nature of CCPs. Classical analysis techniques may provide insights under certain conditions but frequently fail to account for complex time-dependent behaviors (Mahapatra et al., 2014). Moreover, conventional causal analysis is often limited by the lack of mechanistic understanding and baseline data. In order to effectively manage amine emissions and optimize performance of a CCP, advanced modeling and control strategies are required to monitor dynamic operation in energy systems (Li et al., 2016).

Traditional monitoring approaches have focused on physical and chemical interventions. Operators adjust process parameters, such as stripper temperatures and solvent flow rates, to reduce volatile amine losses (Knudsen et al., 2011). Water wash (WW) systems are widely used to capture amine vapors before they exit the absorber, while anti-mist technologies like demisters or enhanced WW sections reduce aerosol-based emissions (Massarweh & Abushaikha, 2024; Zhang et al., 2021). In addition, maintaining solvent quality through the addition of stabilizers or regular solvent reclamation is critical to limiting emissions from degradation products such as nitrosamines. Emission monitoring in traditional setups relies heavily on hardware-based analytical methods such as online Fourier-transform infrared spectroscopy (FTIR) (Giechaskiel & Clairotte, 2021), Ion-molecule reaction-mass spectroscopy (IMR-MS) (Drageset et al., 2022), gas chromatography-mass spectrometry (GC-MS) (Kadadou et al., 2024), and aerosol particle counters (Marina-Montes et al., 2021). Performance of these monitoring techniques generally depends on standard process measurements like pH, $CO_2$ loading, and energy consumption calculations. They are typically manual or rule-based with simple process identifier loops and operator interventions when alarms or thresholds are exceeded. Due to recent advances in computational power and machine learning (ML), it facilitates the development of predictive models to manage amine emissions and optimize system performance in CCPs. ML algorithms such as Linear Regression, Support Vector Machines, and Random Forests have been applied to predict key operational metrics like amine loss and $CO_2$ capture efficiency. However, these models often face limitations in accurately forecasting over longer time horizons due to the complex and nonlinear temporal relationships between process input variables and system output parameters. Deep learning (DL) algorithms, particularly Long Short-Term Memory (LSTM) networks and Convolutional Neural Networks (CNNs) demonstrated improved capabilities in capturing long-term dependencies within time-series data, enabling more accurate multi-hour to multi-day forecasts of plant operations. Despite their strengths, these models are still challenged by issues such as sensor noise, limited data quality, and uncertainty in real-time applications. To address these limitations, researchers are increasingly implementing integrated approaches that combine interpretable ML data processing techniques with advanced DL network frameworks. For example, integrating methods of feature engineering in advanced DL models such as LSTM and CNN, which enhance predictive performance and offer insights into the relative influence of operational input parameters. This approach is critical for enabling more reliable, data-driven decision-making in dynamic and emission-sensitive plant operations.

Recent studies have increasingly explored the use of ML approaches for forecasting carbon captures and overall system performance in CCPs (Fu et al., 2022; Jablonka et al., 2023; Sabeena, 2023). For example, Ashraf et al. employed a machine learning-based approach using Data-Information Integrated Neural Network (DINN) models to predict key performance indicators of fossil-fuel-based power plants (Ashraf, 2024). Specifically, their work focuses on forecasting thermal efficiency, power output, and heat rate by integrating operational data and relevant system information into the neural network framework. Muhammad et al. utilized support vector regression and artificial neural network (ANNs) models to optimize $CO_2$ capture level from the FG in the absorption column is investigated for the post-combustion carbon capture process using MEA (Ashraf & Dua, 2023). The focus of this work has been only on the absorption column, not taking into account other aspects of plant operation, such as the stripper, of the whole carbon capture process. Jablonka et al. developed emission forecasting by using gradient-boosted decision tree and temporal convolution network for effects of emission by the intermittent operation of power plant (Jablonka et al., 2023). Recently, Hosseinifard et al. introduced multiple ML approaches including KNN, logistic regression, Gaussian processes, decision trees, random forest, AdaBoost, gradient boosting, support vector classification, and ANNs to predict and optimize amine solvent selection for a post-combustion carbon system (Hosseinifard et al., 2025). Rapelli et al. applied a LSTM autoencoder network to predict amine emission by using time-series data (Rapelli et al., 2024). However, their approach was constrained by causal impact analysis of input operational variables as a result lack of recommendation to mitigate of the amine emission. Despite these advancements, common limitations across studies include overfitting due to small datasets, poor interpretability of complex models, and difficulty incorporating uncertainty. Further improvements may involve integrating real-time data engineering and assimilation techniques such as the feature engineering methods like lag feature, rolling statistics windows feature, different features, filtering feature etc. to enhance model performance (Wang & Liu, 2022). These enhancements could pave the way for more robust, adaptive, and trustworthy forecasting systems in next-generation CCP operations.

In this study, we explore the concept of real-time forecasting through feature engineering of historical data, a strategy widely utilized in fields such as meteorology, environmental monitoring, stock market prediction, and industrial process control (Gülmez, 2023; Olawade et al., 2024). Feature engineering plays a crucial role in determining the predictive capability of a developed model. Thus, raw sensor data related to amine emissions and system performance must be transformed into more informative and predictive features prior to model training. Techniques like lag feature creation and rolling statistical features can significantly enhance model reliability and accuracy by embedding temporal patterns from observed data. Integrating feature engineering with machine learning enables the development of forecasting models that are both faster and more precise. This approach is relatively recent within the machine learning field and leverages ML's capacity to learn from past experiences of data acquisition processes, often optimized using Bayesian methods (Snoek et al., 2012). Additionally, combining feature analysis with deep learning helps extract high-dimensional features, leading to improved approximations of complex nonlinear systems. Building on these ideas, we developed purely data-driven deep learning forecasting models based on LSTM networks (Sherstinsky, 2020), achieving superior performance and improved computational efficiency.

The goal of this study is to present a framework for integrating ML models to enhance the accuracy and robustness of forecasting for amine emission and plant performance across multiple cases. First, it aims to develop a model capable of real-time prediction of future amine emissions and system performance at various time horizons using historical and current plant data, enabling timely emission control and operational adjustments. Second, it seeks to perform causal impact analysis by generating counterfactual baselines to evaluate the effects of specific operational changes or stress tests on emission levels and performance. Third, the model is used to support emission mitigation through scenario-based simulations (what-if analyses), assessing how interventions in plant operations, such as WW temperatures, solvent temperature etc would affect overall emissions and performance. Recurrent neural networks incorporating LSTM architectures such as vanilla(baseline) LSTM (Hochreiter & Schmidhuber, 1997), bidirectional LSTM (biLSTM) (Graves & Schmidhuber, 2005), stacked LSTM (stackedLSTM) (Jahromi et al., 2020) , and convolutional LSTM (convLSTM) (SHI et al., 2015) were adopted as the core predictive models due to their robust ability to capture sequential dependencies and complex time-series dynamics in CCP operations. We employed feature engineering techniques such as lag features and rolling statistics to improve the learning capability of our ML learning models by creating new inputs that capture temporal patterns in the data, thereby strengthening ML-based forecasts and helping to reduce prediction uncertainty. ML models were specifically trained to forecast daily amine emissions and system performance. Unlike traditional emission forecasting methods that rely solely on mechanistic or statistical models, this ML approach enhances adaptability to operational variability and changing environmental conditions. By combining machine learning's capability to update in real time with its strength in capturing complex nonlinear patterns in amine emissions, this approach supports more responsive amine emission control strategies and enables more effective early warning systems, which are highly critical for environmental protection, and cost-effective CCPs operation.

## 2. Methodology
### 2.1. Data Collection and Preprocessing

In this study, CCP's data were obtained from the U.S. Department of Energy (DOE), the National Energy Technology Laboratory (NETL), which is a test campaign carried out at the Technology Centre Mongstad (TCM) in Norway by the Electric Power Research Institutes, Inc (EPRI), which collected between November 1 and November 23, 2020. TCM was established in 2012, serves as a facility for testing, verifying, and demonstrating various post-combustion $CO_2$ (PCC) capture technologies ("Technology Centre Mongstad | Test Centre for $CO_2$ Capture,"). The test campaign carried out baseline testing of CESAR1 solvent (mixture of 27% wt 2-amino-2-methyl-1-propanol (AMP) and 13% wt piperazine (PZ)) using FG from a nearby combined cycle gas turbine-based heat-and-power (CHP) source. The $CO_2$ concentration was regulated at 5% to replicate conditions typical of advanced gas turbine FG. Subsequent testing involved FG from a residue fluid catalytic cracker (RFCC) source, which features a higher $CO_2$ concentration (refer to **Figure 1** for a process flow diagram when treating RFCC FG). The primary objective of this test campaign was to generate data that supports the reduction of costs and mitigation of technical, environmental, and financial risks associated with the commercial deployment of PCC systems using the CESAR1 solvent. A key focus was establishing real-time operational insights into amine emissions and the CCP's performance.

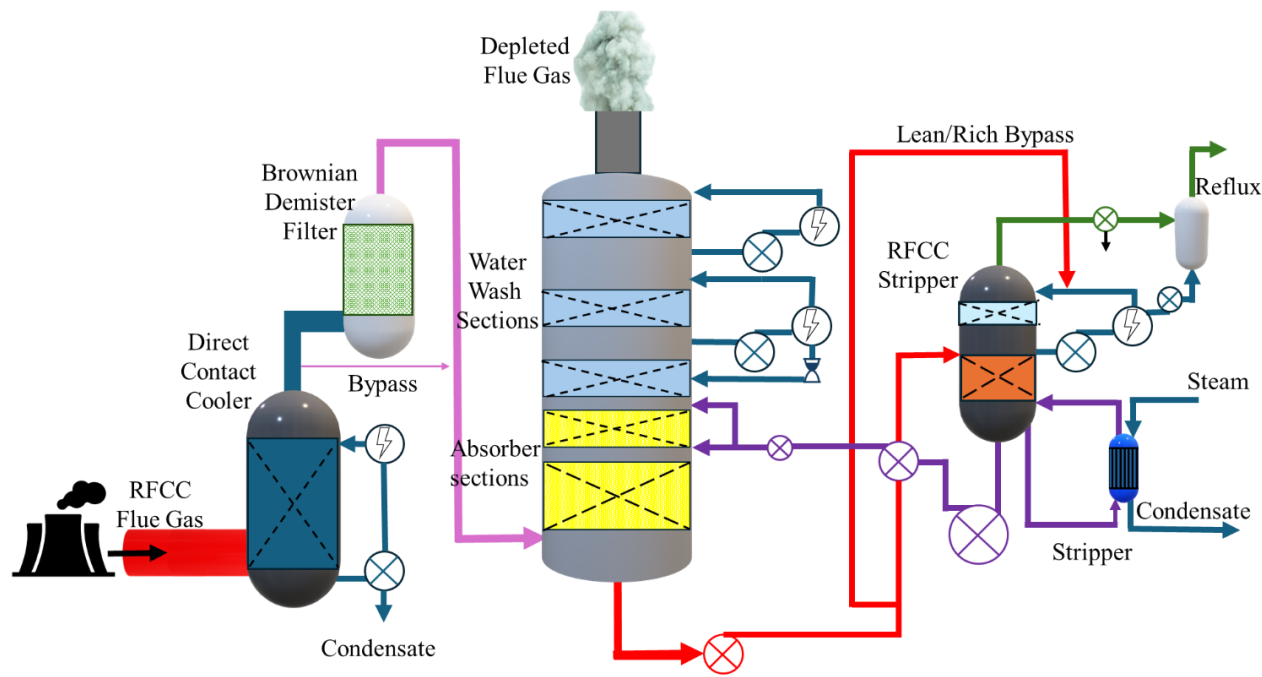

**Figure 1**: Process flow diagram for TCM Amine Plant during EPRI CESAR1 RFCC Testing

The generated datasets record several input and output variables of the plant such as FG inlet properties, system performance, solvent circulation parameters, depleted FG composition, amine emissions WW parameters and $CO_2$ product properties. We received three time-series datasets: the first consists of 10-minute interval collected from November 1 to 15, 2020; the second contains 5-minute interval from November 16 to 23, 2023; and the third includes 5-minute interval collected from November 1 to 15, 2023. For model training, we generated four different scenarios from the received datasets: two with 10-minute intervals covering 15-day and 23-day periods, and two with 5-minute intervals for the same durations. The objective was to compare model performance with respect to sampling interval and the volume of training data. There were some missing values, particularly in the 5-minute interval dataset from November 16 to 23, 2020, which were filled using the time-based interpolation method provided by the Python library. To convert this 5-minute interval data into a 10-minute interval, every alternate (odd-indexed) row was removed, effectively retaining one observation every 10 minutes. After this, we constructed the 23-day dataset, by concatenating the 15-day (November 1–15, 2020) and 8-day (November 16–23, 2020) datasets for both the 10-minute and 5-minute interval cases.

To enhance model robustness, we applied feature engineering techniques, including lag and rolling statistical features. They play a crucial role in time series data analysis and modeling. Lag features help capture temporal dependencies by incorporating past values of a variable as input features, enabling models to learn from historical patterns and autocorrelations. This is particularly important in forecasting tasks where future values are influenced by previous observations. On the other hand, rolling statistics such as rolling mean and standard deviation provide insights into local trends and variability over time. These features help smooth short-term fluctuations and highlight broader patterns, making them valuable for detecting seasonality, shifts, or anomalies in the data. Together, lag and rolling statistical features enhance the predictive power of machine learning models by embedding time-based dynamics directly into the feature set. In this work, we applied 1-hour lag, corresponding to 6-time steps for the 10-minute interval data and 12-time steps for the 5-minute interval data. Furthermore, rolling mean and standard deviation features were computed using window sizes of 30 minutes, 1 hour, 2 hours, and 3 hours to capture short- and medium-term temporal dynamics.

## 2.2. Forecasting of Amine Emissions and System Performance

The machine learning models developed in this study, including the LSTM-based architectures described earlier, utilize historical time-series data to forecast future amine emissions and system performance. Specifically, the model is trained on sequences of past operational data such as 10 minute or 5 minute time interval continuous input variables to predict amine emissions at future horizons up to 2 or 3 days. This is implemented using a sliding window approach, where the input sequence is continuously updated with newly observed data to make the next prediction. While the model is theoretically capable of forecasting across a wide range of future time horizons, predictive accuracy generally declines with increasing forecast lead time. During development, both 10-minute and 5-minute interval datasets were evaluated. The 5-minute interval data consistently outperformed the 10-minute data, so all prediction analyses were conducted using the 5-minute dataset, covering 15-day and 23-day periods. The 23-day dataset achieved better performance than the 15-day dataset, and thus it was used for training and testing the model in the manuscript.

To evaluate model performance, we apply it to test data that is not included during training or validation. However, it is important to note that this validation approach is intentionally conservative. The test dataset comprises baseline operating conditions and intervention scenario represents stepwise changes intended to mimic operational stress that were not present during the training phase. These stress tests decouple prediction points from historical patterns, meaning the model cannot infer future behavior from learned trends alone. Thus, while the predictive accuracy on these unseen cases may appear limited, this evaluation framework realistically reflects how the model might perform under real-world conditions. The use of such rigorous testing highlights the robustness and generalization ability of the developed ML models in anticipating emissions and performance shifts due to variable input conditions.

To develop forecasting models for amine emissions and system performance, we trained the models using four distinct datasets, each differing in temporal resolution and duration. Specifically, we used datasets with 5-minute and 10-minute time intervals, each prepared for two time periods: 15 days and 23 days. This allows us to systematically evaluate the influence of sampling frequency and data length on model accuracy and robustness. For amine emissions forecasting, models were developed specifically for four key emission indicators: AMP measured via FTIR, AMP measured via IMR-MS, Piperazine measured via FTIR, and Piperazine measured via IMR-MS. These outputs capture the concentration and detection method-based variability in amine release under different operational scenarios. For system performance forecasting, we selected critical output variables that reflect the efficiency of the capture system. These include $CO_2$ product flow rate, absorber section outlet temperature, depleted FG outlet temperature, and RFCC stripper bottom temperature. Together, these variables provide a comprehensive view of both emission characteristics and system operational dynamics.

### 2.3. Causal Impact Analysis

The causal impact analysis of amine emissions and system performance involves assessing how specific interventions or variations in input conditions such as changes in operational parameters affect both the emission of amines and the overall efficiency of the system. This type of analysis is particularly valuable in the context of maintaining electricity grid stability, especially when fossil fuel-based power plants are used to compensate for the intermittency of energy demand. The main aspect of this analysis is to understand the operational behavior of carbon capture systems under flexible power plant operations. As power generation fluctuates to match energy demand, it becomes essential to evaluate how the $CO_2$ capture system responds and whether it can maintain environmental regulatory compliance of amine emissions. Therefore, this kind of analysis provides a framework for examining how to optimally operate $CO_2$ capture plants to balance grid demands while minimizing emissions and aligning with future environmental regulations. We investigated the effects of intermittent operational scenarios that may significantly influence amine emissions and the overall performance of carbon capture systems. The baseline condition was defined using test data corresponding to steady-state operation, where the capture plant functions under a standard, constant load. Each intermittent scenario was then designed to simulate variations in power plant load on a daily basis, resulting in corresponding fluctuations in the FG flow rate entering the absorber column. While FG load is a primary factor impacting system dynamics, amine emissions and performance outcomes are also affected by several other process variables within the capture plant. These include operational settings of the WW section, characteristics of the lean solvent (such as temperature and concentration), and the physicochemical properties of the FG, including its composition.

We performed both single-feature and multi-features causal impact analyses to evaluate the effects of input variable changes on amine emission and system performance. We introduced an intervention by changing only one input variable at a time, while keeping all other variables constant during single feature impact analysis. In contrast, the multi-feature analysis involved simultaneous intervention on two input variables while keeping the remaining variables constant. This allows us to assess the combined influence of multiple operational factors and how their interaction affects on the amine emissions and system performance. In this study, eight distinct input variables were utilized to analyze their impact on amine emission and system performance. These input variables include: FG inlet flow rate, FG temperature, lean solvent flow rate, lean solvent temperature, upper WW water flow, upper WW inlet temperature, lower WW water flow, and lower WW inlet temperature. Each of these parameters plays a critical role in the operation of the post-combustion carbon capture system, influencing both the amine emission and plat performance. During the analysis, each input variable was systematically perturbed within a range of ±20% relative to its normal operation. This variation was applied to simulate potential operational deviations and assess the corresponding causal impacts on amine emissions and system performance. By exploring this range, we aimed to capture both conservative and extreme changes that may occur during flexible operation of the capture plant.

## 3. Results and Discussion

### 3.1. Forecasting Analysis

#### 3.1.1. Amine Emission

We built forecasting models for emission of AMP and Piperazine measured by FTIR and IMR-MS techniques. We conducted extensive experimentation with LSTM variants across four datasets, which differed in time intervals (5 and 10 minutes), as outlined in **Section 2.1**. After applying feature engineering, Bayesian optimization, and cross-validation, we evaluated the performance of four LSTM variants: BasicLSTM, BiLSTM, StackedLSTM, and ConvLSTM. The 5-minute interval dataset showed the highest prediction accuracy among the tested configurations. Based on these results, all final forecasting models were trained and evaluated using the 5-minute interval dataset to ensure optimal performance. We also explored forecasting horizons from 1 to 3 days and observed that accuracy decreased as the prediction window increased. In this paper, we present 3-day forecasting results and best-performing models for them are summarized in **Table 1**.

**Table 1**: Summary of best model and their values performance metrices for amine emissions using 5-min interval dataset collected between November 1 and November 23, 2020.

| Amine Emission | Best Model | Model's Parameters | Performance Metrics ||||| 
|---|---|---|---|---|---|---|---|
| | | | MSE | RMSE | MAE | MAPE | $R^2$ |
| AMP FTIR | BiLSTM | 133,761 | 0.00295 | 0.01717 | 0.013565 | 7.9% | 0.70 |
| Piperazine FTIR | StackedLSTM | 50,497 | 0.000202 | 0.01412 | 0.00875 | 13.1% | 0.81 |
| AMP IMR-MS | BiLSTM | 133,761 | 0.000032 | 0.005649 | 0.001533 | 2.5% | 0.97 |
| Piperazine IMR-MS | BasicLSTM | 50,497 | 0.000033 | 0.005755 | 0.001858 | 2.7% | 0.96 |

**Table 1** shows that our forecasting models deliver strong predictive performance across all four measured amine emissions. The BiLSTM model performed best for AMP emissions measured by both FTIR and IMR-MS. For AMP FTIR, BiLSTM achieved an MSE of 0.00295, RMSE of 0.01717, MAE of 0.01357, MAPE of 7.9%, and an $R^2$ of 0.70, indicating solid accuracy with moderate variance explained. For AMP IMR-MS, BiLSTM performed even better, with a very low MSE of 0.000032 and a high $R^2$ of 0.97, showing excellent agreement between predicted and actual values. This is also clearly visible in **Figure 2**, where the predicted values highly aligned with the true emission data. For Piperazine measured by FTIR, the Stacked LSTM model gave the best results, achieving an $R^2$ of 0.81 and an MSE of 0.000202. Although the MAE was low at 0.00875, the MAPE was relatively high at 13.1%, likely due to underprediction of sudden emission spikes. In the case of Piperazine IMR-MS, the BasicLSTM model performed best, with a strong $R^2$ of 0.96 and low error values (MSE: 0.000033, RMSE: 0.005755). The visual comparisons in **Figure 2** confirm the high alignment between the predicted and true values.

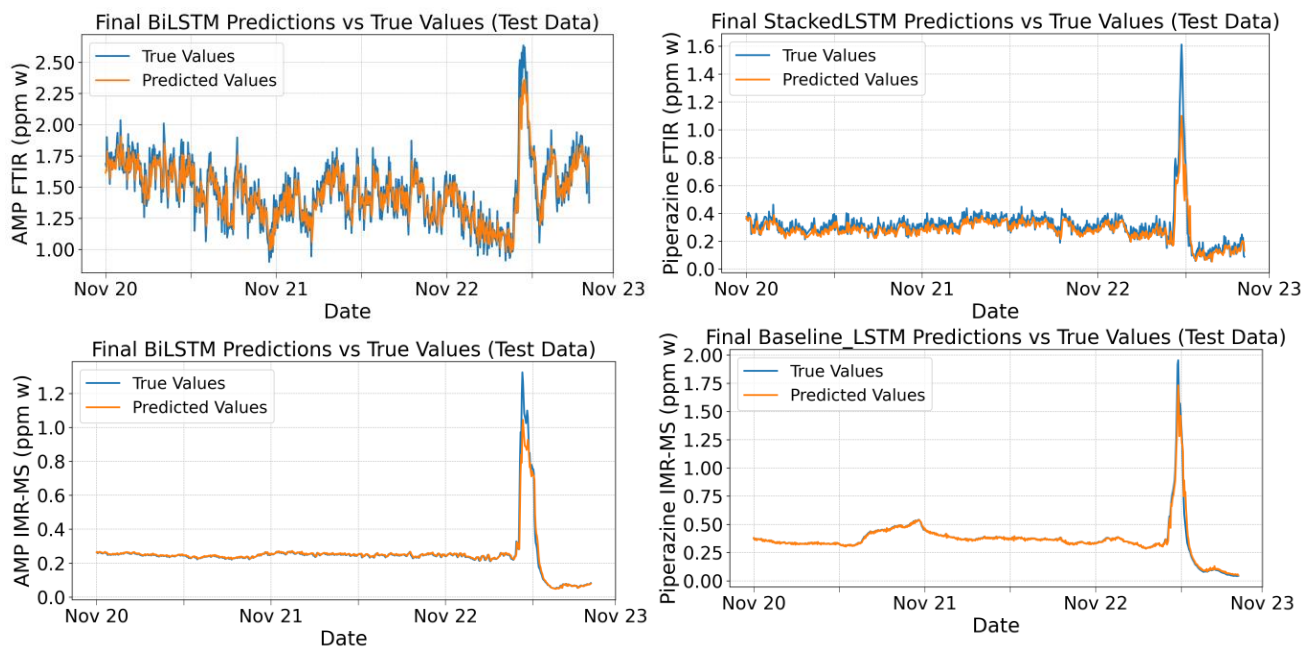

**Figure 2:** Plot of the predicted versus actual amine emission values over a continuous 3-day period, generated using the best-performing LSTM model trained on a 23-day emissions dataset; AMP FTIR (top left), Piperazine FTIR (top right), AMP IMR-MS (bottom left), and Piperazine IMR-MS (bottom right)

These results highlight that choosing the right LSTM architecture such as BiLSTM for AMP across different measurement techniques and StackedLSTM for Piperazine FTIR or BasicLSTM for Piperazine IMR-MS can lead to accurate and reliable forecasting. Moreover, **Figure 2** illustrates that our models not only closely align the overall emission trends but also effectively capture sharp fluctuations over a 3-day forecasting horizon, indicating their robustness and precision. Therefore, our results demonstrate that the predicted values for AMP and Piperazine emissions closely align with the actual measurements. Our modeling approach accurately captures temporal patterns and variations, including both gradual trends and sudden fluctuations. This indicates that the proposed LSTM-based forecasting framework is highly effective and reliable for modeling amine emissions with strong predictive precision.

### 3.1.2. System Performance

We use same modeling approach applied to amine emissions to develop forecasting models for key system performance variables: $CO_2$ product flow, absorber outlet temperature before WW, depleted FG outlet temperature, and RFCC stripper bottom temperature. As before, four LSTM variants such as BasicLSTM, BiLSTM, StackedLSTM, and ConvLSTM were trained and evaluated using the 5-minute interval dataset. **Table 2** summarizes the best-performing models for each parameter based on standard performance metrics.

**Table 2**: Summary of best model and their values performance metrices for system performance using 5-min interval dataset collected between November 1 and November 23, 2020.

| System Performance | Best Model | Model's Parameters | Performance Metrics ||||| 
|---|---|---|---|---|---|---|---|
| | | | MSE | RMSE | MAE | MAPE | $R^2$ |
| CO2 Product flow | BiLSTM | 133,761 | 0.00016 | 0.01267 | 0.01267 | 0.43% | 0.99 |
| Abs outlet Temp Before WW | BiLSTM | 133,761 | 0.00030 | 0.01739 | 0.00700 | 0.72% | 0.99 |
| Depleted FG outlet Temp | StackedLSTM | 50,497 | 0.00018 | 0.01326 | 0.00589 | 1.37% | 0.99 |
| RFCC stripper Bottom Temp | BiLSTM | 133,761 | 0.00039 | 0.01968 | 0.00598 | 0.36% | 0.99 |

**Table 2** illustrates that the BiLSTM model produced the best performance for three out of the four variables. For $CO_2$ product flow, the BiLSTM model showed the highest accuracy, achieving a very low MSE of 0.00016, RMSE of 0.01267, MAE of 0.01267, MAPE of just 0.43%, and an excellent R² of 0.99, indicating near-perfect prediction. Similarly, for the absorber outlet temperature before the WW, BiLSTM again performed best, with an MSE of 0.00030, RMSE of 0.01739, MAE of 0.00700, MAPE of 0.72%, and R² of 0.99, demonstrating its effectiveness in modeling temperature dynamics. The depleted FG outlet temperature was best predicted by the StackedLSTM model, which achieved an MSE of 0.00018, RMSE of 0.01326, MAE of 0.00589, MAPE of 1.37%, and R² of 0.99, demonstrating excellent fit. For the RFCC stripper bottom temperature, the BiLSTM model again provided the most accurate results, with an MSE of 0.00039, RMSE of 0.01968, MAE of 0.00598, MAPE of 0.36%, and an R² of 0.99, showing accurate prediction.

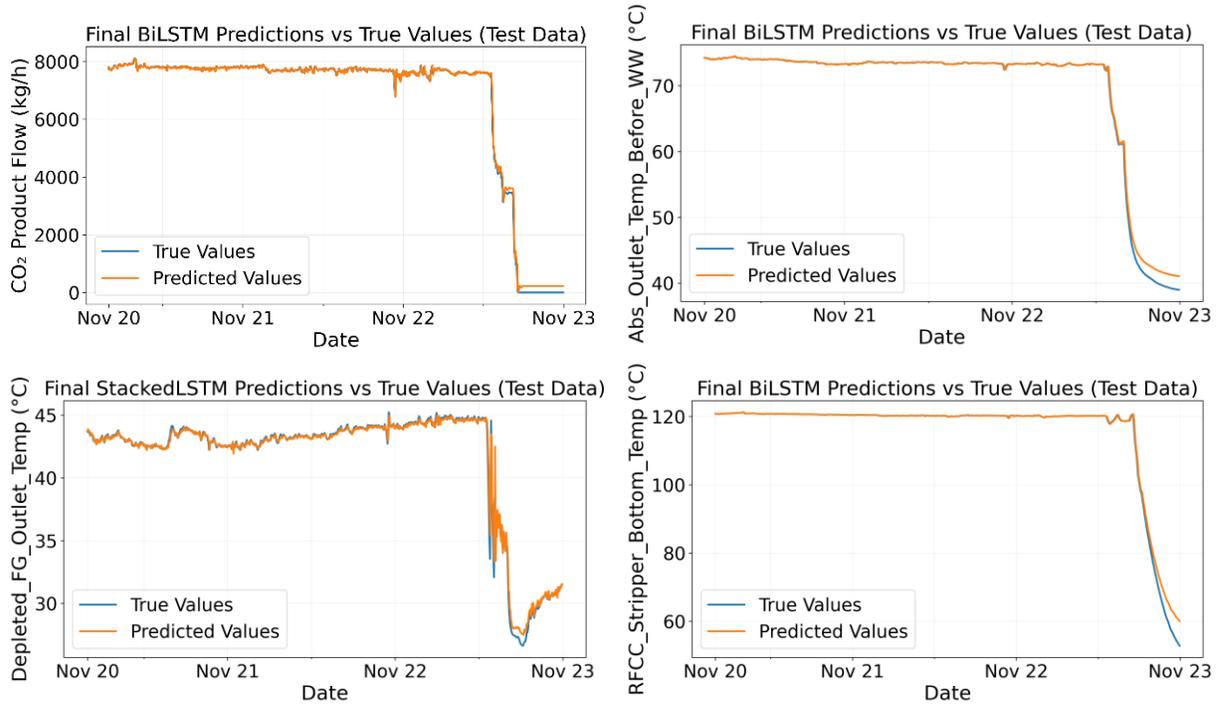

**Figure 3:** Plot of the predicted versus actual system performance values over a continuous 3-day period, generated using the best-performing LSTM model trained on a 23-day emissions dataset; $CO_2$ product flow (top left), Absorber outlet temperature before water wash (top right), Depleted flue gas outlet temperature (bottom left), and RFCC stripper bottom temp (bottom right)

These results highlight the strong generalization ability of LSTM-based models for time-series forecasting across different key system performance variables. The consistently high $R^2$ values and low error metrics confirm that these models can reliably capture both the trends and fluctuations of key performance indicators in the system. **Figure 3** visually supports these findings, showing predicted values tracking closely with true values across all variables for the 3-day forecasting window. Notably, even abrupt changes in system behavior were captured accurately by the models, reflecting their robustness and ability to generalize across different operational dynamics.

### 3.2. Causal Impact Analysis

The key objective of impact analysis is to estimate and mitigate the effects of intermittent operational scenarios that can substantially impact amine emissions and overall system performance. These scenarios are driven by fluctuations in power generation resulting from variable demand in fuel-based energy systems. We focused on intervention scenarios of input variables of the system that could potentially have strong effect on amine emission and system performance. To evaluate causal impact, it is essential to establish a baseline or normal operational input feature profile. This baseline represents as a reference point, enabling the simulation of intervention scenarios and the estimation of their corresponding impacts on system behavior. In this study, the provided dataset is assumed to represent baseline operational conditions corresponding to normal power plant operations. The best-performing models developed in the previous step (**Section 3.1**) were used as baseline models for predicting amine emissions and system performance parameters. These models served as the reference framework for evaluating the effects of input interventions during causal impact analysis. Accordingly, each scenario is evaluated by examining deviations in daily system behavior, reflecting variations in input features that result from differing operational conditions.

For the causal impact analysis, we utilized the 5-minute interval dataset spanning 15 days, collected during the initial operational period at TCM. Dataset provides sufficient temporal resolution and consistency for conducting a robust causal impact analysis. We employed eight key plant input features: FG inlet flow, FG inlet temperature, lean solvent flow, lean solvent temperature, upper WW water inlet flow, upper WW water inlet temperature, lower WW water inlet flow, and lower WW water inlet temperature, to evaluate their impact on amine emissions and system performance. This analysis was performed by applying controlled interventions to these inputs, varying each from –20% to +20% in 5% increments relative to baseline values. Both single-feature and multi/two-feature interventions were estimated, while all other features were held constant at their baseline levels. Scenario modeling conducted in this study is based on several key assumptions: (1) The underlying dynamics of the system do not change during the intervention period. (2) Other potentially correlated input variables remain unchanged in both single-feature and multi-feature causal impact analysis, (3) The average percentage change in amine emissions and system performance parameters in 2-day forecasting period is a valid and meaningful indicator of the overall impact of the interventions.

### 3.2.1. Single-Feature Analysis of Amine Emissions

We performed causal impact analysis using single-feature interventions applied to individual input variables. The analysis was performed for both amines, AMP and Piperazine, measured using FTIR and IMR-MS techniques, each offering distinct insights into causal relationships. A summary of the results is presented as a heatmap in **Figure 4**, which illustrates the varying degrees of causal impact across different operational parameters and intervention magnitudes. The color-coded matrix illustrates the average percentage difference in predicted emissions as a function of varying key operational input parameters within a ±20% perturbation range (in 5% increments) from the baseline. The observed patterns highlight how specific input variations influence amine emissions under different measurement methods. It reveals critical sensitivities and nonlinear relationships between plant input features and their impact on amine emissions.

**Figure 4** illustrates that the emissions of AMP and piperazine, as measured using FTIR and IMR-MS, show significant sensitivity to key operational parameters. Among these, lean solvent temperature, FG inlet flow rate, and WW settings exert the most pronounced effects. For AMP FTIR, the absorber's lean solvent temperature emerges as the dominant variable: a 20% reduction in temperature leads to a 35.2% increase in emissions, whereas a 20% temperature increase results in a 20.1% reduction. This inverse correlation implies that elevated lean solvent temperatures may enhance solvent regeneration or mitigate condensation-related emission mechanisms. Similarly, a 20% decrease in FG inlet flow leads to a 32.6% rise in AMP emissions, potentially due to diminished turbulence or extended gas–liquid contact times. Adjustments to upper WW flow also affect AMP release, with a 20% increase in flow achieving up to a 22.2% emission reduction. In contrast, increasing upper WW water temperature by 20% elevates emissions by as much as 16.1%, likely due to reduced cooling efficacy. Lower-stage wash water temperature exerts minimal influence, while increases in lower-stage WW flow yield only modest emission reductions. Moreover, both lean solvent flow and FG temperature show limited impact, though a slight decrease in emissions is observed with higher FG temperature. As a demonstration.

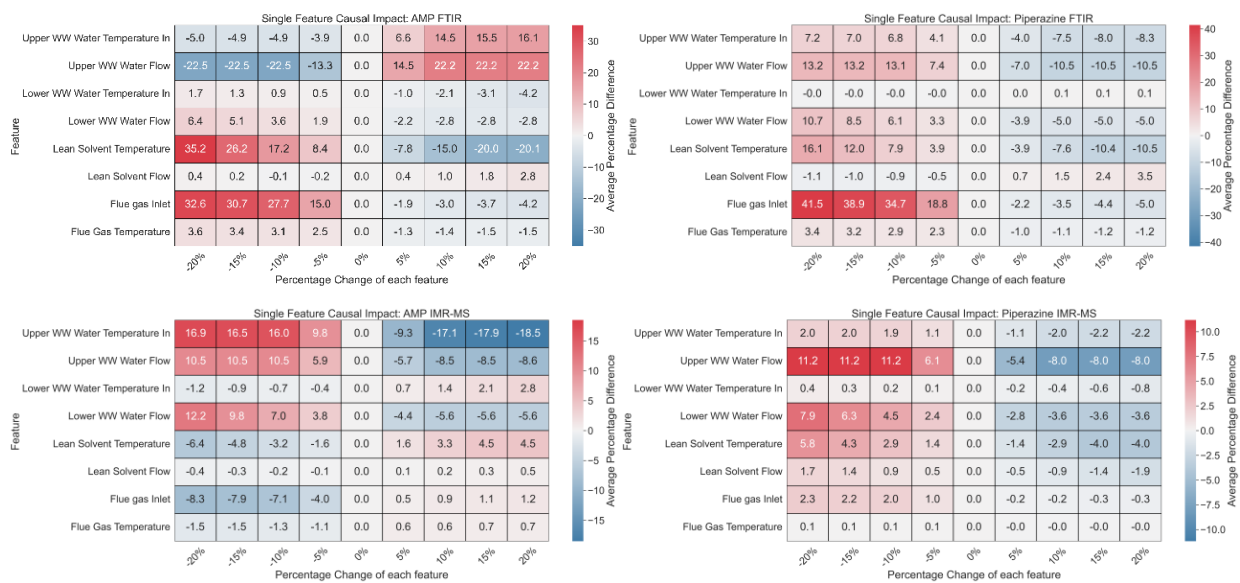

**Figure 4:** Heatmaps illustrating the causal impact analysis of amine emissions: AMP FTIR (top left), Piperazine FTIR (top right), AMP IMR-MS (bottom left), and Piperazine IMR-MS (bottom right) with respect to plant input features. The X-axis represents the percentage change in each individual input feature from the baseline, while all other features are held constant at baseline. The Y-axis lists the plant input features, and each cell in the matrix shows the corresponding average percentage change in amine emissions.

For piperazine FTIR, the FG inlet flow rate is identified as the most influential operational variable. A 20% reduction in FG flow results in a substantial 41.5% increase in emissions, indicating a strong sensitivity even to minor reductions in flow. Lean solvent temperature also exhibits a notable inverse relationship with emissions: a 20% increase in temperature reduces emissions by 10.5%, whereas a 20% decrease leads to a 16.1% increase, suggesting that elevated solvent temperatures enhance desorption efficiency. Upper WW parameters significantly affect emission behavior. A 20% reduction in upper WW flow leads to an emission decrease of up to 13.2%, while a 20% increase in flow corresponds to a 10.5% reduction in emissions, reflecting the role of liquid distribution and surface renewal in mitigating volatile amine release. In contrast, lower WW flow exhibits a nonlinear and more complex relationship with piperazine emissions, while its temperature shows negligible impact. Finally, both lean solvent flow and FG temperature exert only minor influence, with resulting emission variations limited to approximately 1–3%, depending on the direction of the operational change.

In the AMP IMR-MS, upper WW temperature emerges as the most influential variable. A 20% decrease in upper WW temperature leads to an 18.5% reduction in emissions, whereas a 20% increase results in a 16.9% rise, underscoring a pronounced sensitivity to thermal conditions in this section of the system. Lower WW flow also exhibits a notable effect: a 20% decrease in flow is associated with a 12.2% increase in emissions, potentially due to compromised amine capture efficiency at reduced flow rates. Conversely, increasing upper WW flow results in decreased AMP emissions, emphasizing the role of optimized flow conditions in mitigating volatile losses. Interestingly, the influence of lean solvent temperature diverges from FTIR-based observations; in the IMR-MS measurements, lower solvent temperatures correlate with reduced emissions. This inverse trend may reflect differences in volatility dynamics or methodological sensitivities between the two detection techniques. FG flow and lean solvent flow demonstrate only minor effects, with slight reductions in emissions observed when FG flow is decreased, and negligible variation associated with changes in solvent flow. FG temperature contributes marginally to emission behavior, with positive correlation in AMP emission by temperatures perturbation.

For piperazine IMR-MS, upper and lower WW flow rates as the most influential operational variables, with particularly strong effects observed in the upper WW section. A 20% reduction in upper WW flow results in an 11.2% increase in emissions, whereas a 20% increase in flow yields an 8.0% reduction, demonstrating the efficacy of enhanced liquid flow in mitigating amine release. Lower WW flow also impacts emissions: decreasing it by 20% leads to a 7.9% rise in emissions, while a 20% increase produces a 3.6% reduction. Lean solvent temperature and flow exhibit moderate influences on piperazine emissions, suggesting their potential role in fine-tuning system performance. In contrast, FG inlet flow and temperature have negligible effects, indicating that emission control in this context is more dependent on solvent and WW management than on upstream gas-phase conditions. Thus, these results provide a robust foundation for guiding emission control in post-combustion carbon capture processes and point to the need for further exploration of parameter interactions for enhanced process control.

### 3.2.2. Single-Feature Analysis of System Performance

Here, we use the same approach applied to amine emissions to estimate causal impact analysis for key system performance variables: $CO_2$ product flow, absorber outlet temperature before WW, depleted FG outlet temperature, and RFCC stripper bottom temperature. The results are summarized as a heatmap plot and shown in **Figure 5**, which illustrates the varying degrees of impact across different input parameters and intervention magnitudes. The color-coded matrix illustrates the average percentage difference in estimated system performance parameter as a function of varying key operational input parameters within a ±20% perturbation range (in 5% increments) from the baseline.

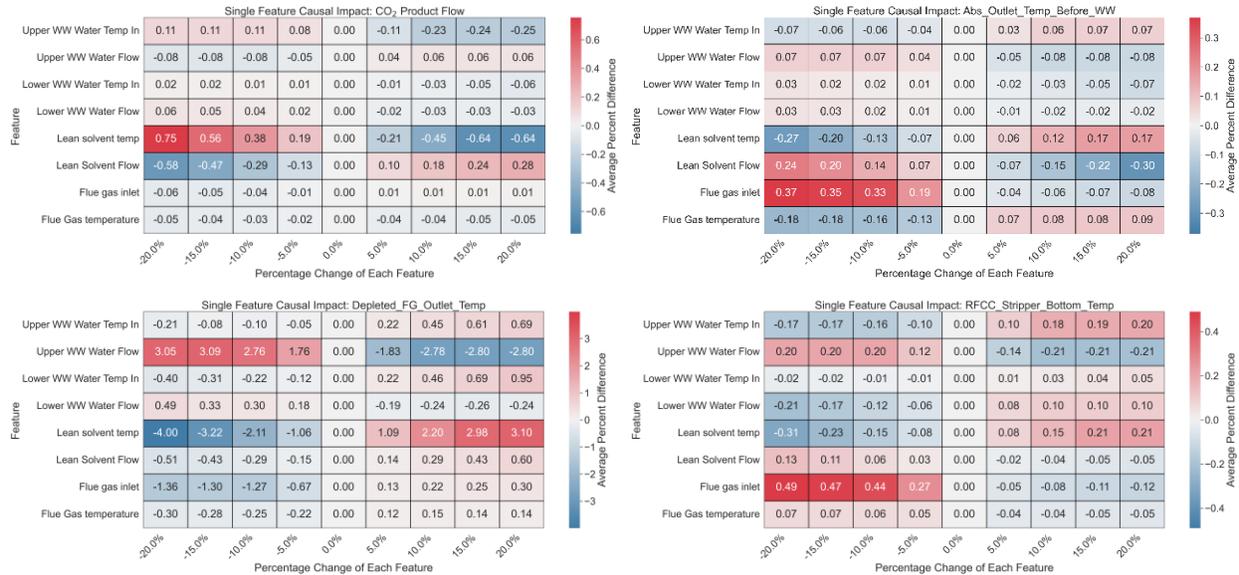

**Figure 5:** Heatmaps illustrating the causal impact analysis of system performance parameters: $CO_2$ product flow (top left), Absorber outlet temperature before water wash (top right), Depleted flue gas outlet temperature (bottom left), and RFCC stripper bottom temp (bottom right) with respect to plant input feature. The X-axis represents the percentage change in each individual input feature from the baseline, while all other features are held constant at baseline. The Y-axis lists the plant input features, and each cell in the matrix shows the corresponding average percentage change in system performance.

**Figure 5** highlights lean solvent temperature as the most influential operational parameter affecting $CO_2$ product flow, showing a strong causal relationship. A 20.0% reduction in lean solvent temperature results in a maximum average increase in $CO_2$ flow of +0.75%, while a 20.0% increase leads to a decrease of –0.64%. This clearly demonstrates the critical role of thermal conditions in optimizing solvent absorption efficiency. In contrast, lean solvent flow exerts a predominantly negative influence, with an average impact of –0.58%. However, at higher perturbation levels, the effect becomes slightly positive (+0.28%), suggesting an initial dilution effect that diminishes as increased circulation compensates through improved mass transfer or solvent availability. Upper WW temperature exhibits an inverse relationship, with a 20.0% reduction producing a moderate average increase in $CO_2$ flow (+0.11%) and a 20.0% increase yielding a decline (–0.25%). This suggests a temperature-driven trade-off between condensation and solvent efficacy. Meanwhile, upper WW flow shows a minor and nonlinear positive effect, with average changes of –0.08% and +0.06% at –20.0% and +20.0% perturbations, respectively, potentially reflecting hydraulic resistance or solvent dilution dynamics. Lower WW temperature and flow offer small but consistent positive contributions, averaging +0.02% and +0.06%, respectively, under 20.0% reductions. However, their impact diminishes at greater deviations, indicating a more limited role in driving $CO_2$ flow changes. FG inlet flow and temperature demonstrate minimal sensitivity, with changes of up to –0.06% and –0.05% under 20.0% reductions, reflecting a robust absorber design that insulates $CO_2$ production from upstream variability. Overall, these findings underscore the dominant influence of lean solvent temperature and to a lesser extent, lean solvent flow on $CO_2$ product flow. While wash water parameters provide secondary stabilization, optimal control of thermal conditions remains essential for maximizing capture efficiency in post-combustion carbon capture systems.

For the absorber outlet temperature measured just before the WW section, FG inlet flow emerges as the most influential parameter, with a positive average effect of +0.37% when reduced by 20.0%. This indicates its significant role as a thermal energy input to the absorber. In contrast, FG temperature exerts a negative average effect of –0.18% under a 20.0% reduction, improving to a modest positive effect of +0.09% when increased by 20.0%. Lean solvent temperature contributes a notable influence, with an average decrease of –0.27% at –20.0%, shifting to a positive effect of +0.17% at +20.0%. This may reflect a transition from a net cooling influence at lower temperatures to thermal stabilization at elevated temperatures. Lean solvent flow demonstrates a nonlinear behavior, yielding a positive average effect of +0.24% at low perturbation levels but declining to –0.30% at higher flow rates, suggesting an initial thermal gain followed by reduced efficiency due to over-dilution or reduced residence time. Upper WW flow shows a minor yet directionally dependent impact: a small positive change (+0.07%) at –20.0% and a slight negative effect (–0.08%) at +20.0%. Upper WW temperature shows a symmetrical and consistently mild impact, with a ±0.07% change observed at both ±20.0% variations. Lower WW temperature and flow exert minimal influence, both showing a slight negative correlation with absorber outlet temperature, reinforcing their secondary role in heat regulation. Thus, absorber outlet temperature is primarily governed by FG flow and lean solvent thermal inputs, with WW parameters offering limited yet stabilizing contributions. Effective management of these variables is essential for maintaining thermal equilibrium and absorber performance.

The heatmap analysis of depleted FG outlet temperature reveals substantial variability driven by key operational parameters. Upper WW flow emerges as a dominant factor, showing a distinct inverse response: a 20.0% reduction results in an average temperature increase of +3.05%, while a 20.0% increase leads to a decrease of –2.80%. This suggests strong thermal coupling, likely due to enhanced or reduced heat removal in the upper WW section. Lean solvent temperature also shows a significant impact, with a strong positive correlation to depleted FG outlet temperature. A 20.0% decrease in solvent temperature leads to an average reduction of –4.0%, while a 20.0% increase raises the outlet temperature by +3.10%. This behavior is consistent with solvent cooling dynamics influencing the absorber's heat profile and gas–liquid equilibrium. Lower WW temperature and flow contribute to minor but consistent positive correlations, reflecting their limited role in thermal moderation downstream. Lean solvent flow similarly displays a positive trend, with an average temperature decrease of –0.51% at –20.0% and a rise of +0.60% at +20.0%, indicating a moderate influence likely tied to circulation rate and heat load distribution. FG inlet flow and temperature also contribute positively, though to a lesser extent. At baseline, a 20.0% reduction in FG inlet flow results in an average decrease of –1.36%, which rises to +0.30% with a 20.0% increase. FG temperature exhibits smaller changes, from –0.30% to +0.15%, suggesting its role is secondary but non-negligible. These results emphasize that upper WW flow and lean solvent temperature are the key control variables for managing depleted FG outlet temperature, while additional adjustments to solvent and WW parameters offer complementary thermal regulation. Strategic coordination of these inputs is essential for maintaining process stability and optimizing energy efficiency in carbon capture operations.

The causal analysis of RFCC stripper bottom temperature identifies FG inlet flow as the most influential variable, exhibiting a notable positive impact of +0.49% when reduced by 20.0%. This confirms its role as a primary thermal input to the stripper. FG temperature, in contrast, shows a modest negative correlation, indicating a less direct but still measurable effect on thermal conditions within the stripper. Lean solvent temperature displays a non-linear positive relationship with stripper bottom temperature. A 20.0% reduction results in an average decrease of –0.31%, while a 20.0% increase leads to a positive shift of +0.21%. This trend suggests an initial cooling effect at lower temperatures, followed by thermal stabilization or recovery at elevated conditions. Lean solvent flow similarly demonstrates a variable response: a 20.0% reduction induces a +0.13% increase in bottom temperature, whereas a 20.0% increase causes a slight reduction (–0.05%), likely due to over-circulation effects such as reduced residence time or increased dilution. Upper WW water flow exhibits a bidirectional influence, with a +0.20% increase in bottom temperature at reduced flow (–20.0%), but a decline of –0.21% at elevated flow (+20.0%). This suggests a flow threshold beyond which cooling becomes excessive or disrupts thermal balance. Upper WW temperature shows a generally positive correlation, shifting from –0.17% at lower temperatures to +0.20% at higher values, possibly due to increased heat retention or reduced condensation. Lower WW flow and temperature both contribute positively to bottom temperature, though the effect of flow is more pronounced. This indicates that lower section wash water flow can support thermal stability, potentially by maintaining optimal hydraulic conditions or enhancing mass transfer efficiency. Overall, these findings highlight the dominant role of FG inlet flow and lean solvent thermal properties in regulating stripper bottom temperature. Wash water parameters, particularly in the upper section, offer additional fine-tuning capabilities to maintain optimal operational stability and thermal performance

### 3.3. Emission Mitigation and Sensitivity Analysis

The causal impact analysis of single-feature intervention provides valuable insights into the significance and magnitude of individual changes applied to the plant. However, during stress operations, multiple parameters are often implicitly altered. By leveraging our predictive model, we can analyze these complex operational data to identify which specific changes in plant operation are most effective in reducing overall amine emissions and enhancing system performance under stress conditions. In this study, we conducted multi-feature causal impact analysis by selecting two input features of different plant operational scenarios.

### 3.3.1. Multi-Feature Analysis of Amine Emissions

We conducted a causal impact analysis using multi-feature interventions by selecting two input variables for each amine of AMP and Piperazine measured by FTIR and IMR-MS methods. The results are visualized as heatmaps, as shown in **Figure 6**, which represents the intervention effects across two key operational features: upper WW water temperature and lean solvent temperature, with intervention applied as percentage deviations. The color-coded matrix displays the average percentage change in predicted emissions resulting with changing input feature within a ±20% range, varied in 5% increments from the baseline.

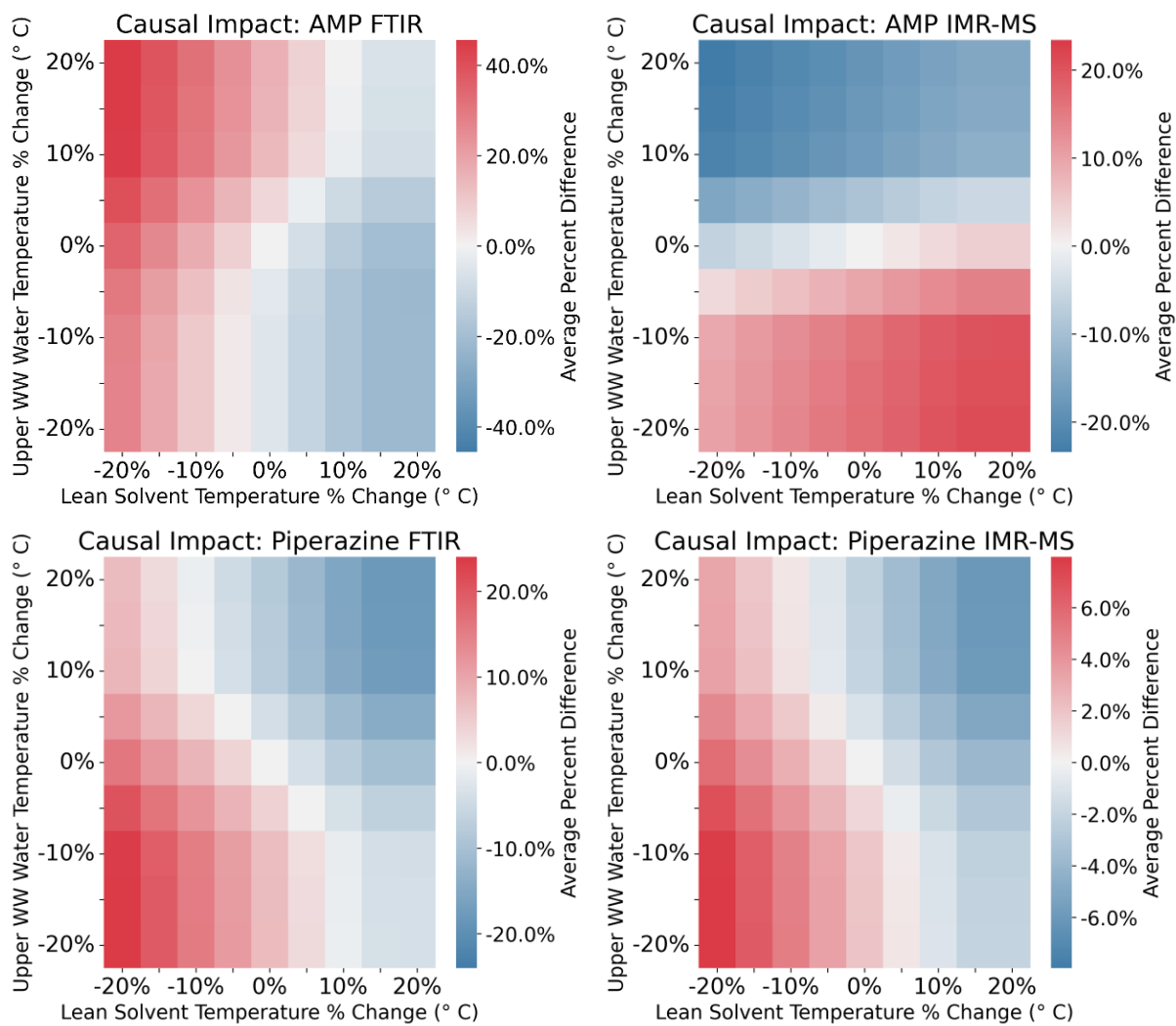

**Figure 6:** Heatmaps illustrating the causal impact analysis of amine emissions; AMP FTIR (top left), Piperazine FTIR (top right), AMP IMR-MS (bottom left), and Piperazine IMR-MS (bottom right) with respect to plant's two input features of *Upper WW Water Temperature* and *Lean Solvent Temperature*. The X-axis and Y-axis represent the percentage change in input features from the baseline, while all other features are held constant at baseline. Each cell in the matrix shows the corresponding average percentage change in specific amine emission.

**Figure 6** illustrates a quantifiable interaction between lean solvent temperature and upper WW temperature in influencing AMP and piperazine emissions measured in both measured FTIR and IMR-MS. For AMP FTIR, this relationship is particularly prominent, with emission changes ranging from –40.0% to +40.0%. Specifically, combinations of lower lean solvent temperatures (–10% to –20%) and elevated upper WW temperatures (+10% to +20%) lead to substantial increases in emissions up to +40.0%, likely due to reduced absorption efficiency and hindered solvent regeneration. As the input temperatures approach their nominal values, the system demonstrates thermal stabilization, and when the lean solvent temperature is increased (+10% to +20%) while the upper WW temperature is decreased (–10% to –20%), a strong emission reduction is observed, reaching up to –40.0%. Interestingly, when both parameters vary in the same direction (both increase or decrease), the resulting changes in emissions are relatively minor, suggesting partial cancellation of thermal effects. A similar interaction pattern is evident in the AMP IMR-MS, though with a reduced magnitude, ranging from –20.0% to +20.0%. This consistency across measurement techniques suggests shared underlying thermodynamic behavior, albeit with differing sensitivity or instrument calibration. As with FTIR, the greatest emission impacts occur when the two temperature variables shift in opposite directions: increasing lean solvent temperature while decreasing upper WW temperature results in elevated emissions, whereas the reverse scenario (lower lean solvent temperature and higher WW temperature) reduces emissions. These findings reinforce the conclusion that amine emissions, particularly for AMP which is governed by the thermodynamic properties of the solvent system, and that careful coordination of lean solvent and upper WW temperatures is critical for minimizing emissions and enhancing absorber performance.

The piperazine FTIR results are closely similar to the AMP IMR-MS findings, showing emission variations within a range of –20.0% to +20.0% in response to ±20% perturbations in lean solvent and upper WW temperatures. The combined influence of these variables reveals a clear negative correlation with piperazine emissions: simultaneous increases in both input temperatures lead to emission reductions, while simultaneous decreases result in increased emissions. This indicates that piperazine FTIR is similar responsive to thermal conditions across both process variables. Piperazine IMR-MS display a more constrained response, with a change range of –6.0% to +6.0%. Despite the smaller magnitude, the directional trend remains consistent with the FTIR results. This reduced sensitivity may stem from either the lower responsiveness of piperazine under IMR-MS detection or intrinsic differences in its thermophysical properties compared to AMP. Overall, a consistent trend emerges emissions tend to decrease at higher lean solvent temperatures and increase at lower ones, with near-neutral behavior around baseline conditions. An exception is noted in AMP FTIR data, where the inverse effect occurs, highlighting a potential interaction between detection method and amine volatility. A mirrored pattern is observed for changes in upper WW temperature, further emphasizing the bidirectional influence of these thermal parameters. These observations underscore that lean solvent and upper WW temperatures are dominant operational features in controlling amine emissions. Their coordinated management is critical for achieving emission minimization and thermal stability in post-combustion carbon capture systems.

### 3.3.2. Multi-Feature Analysis System Performance

**Figure 7** illustrates a measurable interaction between lean solvent temperature and upper WW temperature in influencing key system performance parameters. **Figure 7** demonstrates that the causal impact of upper WW temperature and lean solvent temperature (varied within ±20%) on $CO_2$ product flow result in a negative correlation. Specifically, $CO_2$ flow exhibits an average percentage change between –0.8% and +0.8%, with production increasing as either one input temperature decreases. Notably, reducing upper WW temperature while maintaining constant lean solvent temperature results in increased $CO_2$ flow, whereas increasing lean solvent temperature under constant WW temperature leads to a decline in $CO_2$ production. This interaction pattern underscores the importance of coordinated thermal management between the upper WW and lean solvent streams to optimize $CO_2$ capture efficiency and system performance. Similarly, the absorber outlet temperature exhibits a comparable but less pronounced trend. The average variation ranges from –0.3% to +0.3% over the same ±20% input range. A positive correlation with input temperatures is observed, indicating that increases in either upper WW or lean solvent temperature elevate the absorber outlet temperature.

For the depleted FG outlet temperature, the analysis reveals a pronounced sensitivity to variations in both upper WW temperature and lean solvent temperature, with impacts ranging from –4.0% to +4.0% across a ±20% perturbation range. A positive correlation is observed with both temperature inputs, indicating that elevated upper WW and lean solvent temperatures drive corresponding increases in the depleted FG outlet temperature. It shows that these two parameters are critical for controlling thermal conditions at the absorber outlet. In contrast, the RFCC stripper bottom temperature demonstrates a narrower but consistent response to these same variables, with changes confined to the range of –0.4% to +0.4%. This suggests that while the influence is more moderate, upper WW and lean solvent temperatures still exert a measurable effect on stripper thermal behavior. Overall, the results across all four key performance indicators underscore the central role of upper WW temperature and lean solvent temperature in maintaining system-wide performance. The consistent influence of these parameters highlights the need for precise and coordinated control, particularly of upper WW temperature, in conjunction with lean solvent temperature, to enhance efficiency, minimize thermal deviations, and ensure robust plant operation.

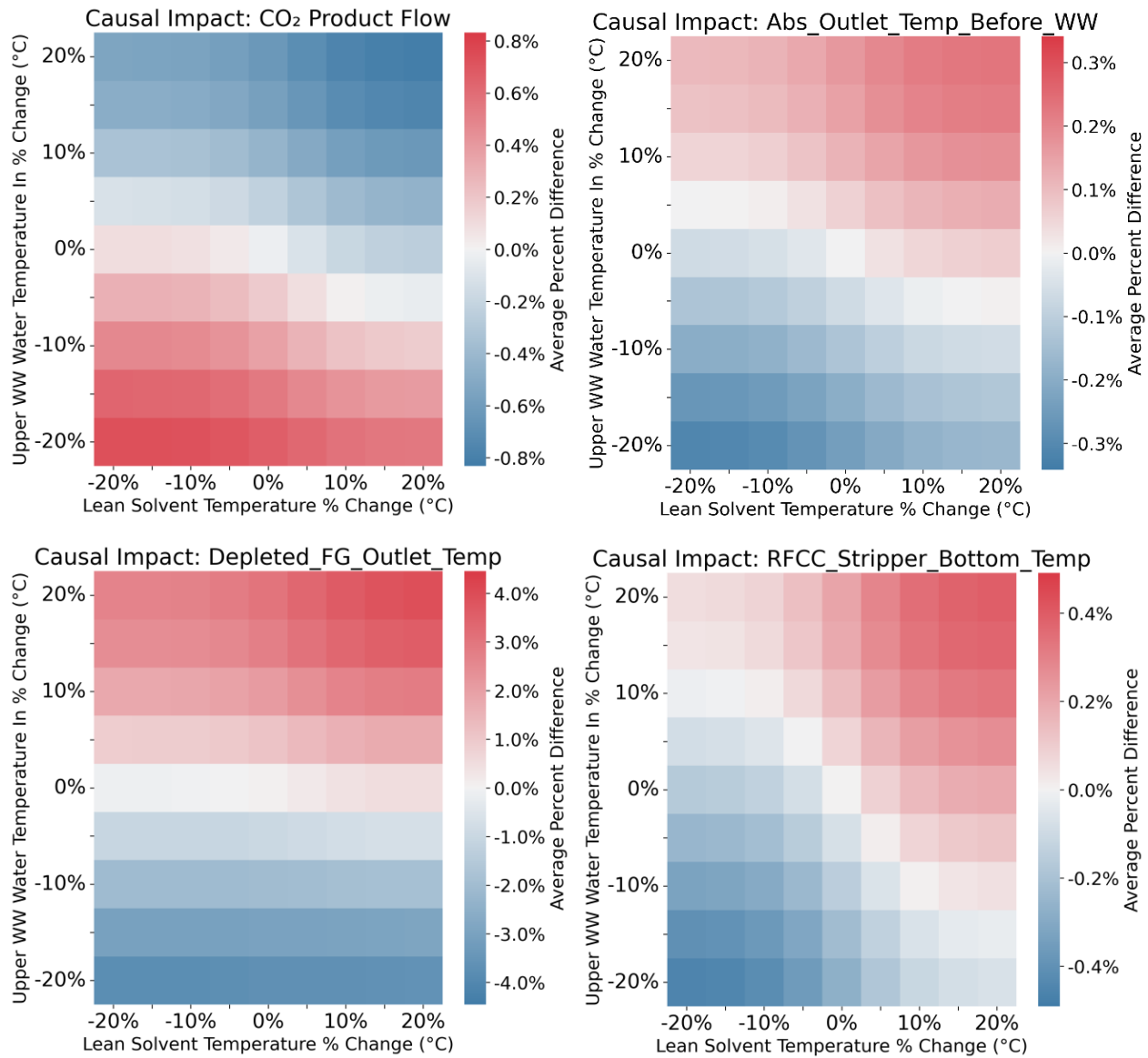

**Figure 7:** Heatmaps illustrating the causal impact analysis of system parameters; CO$_2$ product flow (top left), Absorber outlet temperature before water wash (top right), Depleted flue gas outlet temperature (bottom left), and RFCC stripper bottom temp (bottom right) with respect to plant's two input features of *Upper WW Water Temperature In* and *Lean Solvent Temperature*. The X-axis and Y-axis represent the percentage change in input features from the baseline, while all other features are held constant at baseline. Each cell in the matrix shows the corresponding average percentage change in specific system performance.

## 4. Conclusion and Future Study

We developed a data-driven ML framework to forecast amine emissions and key system performance parameters in an amine-based post-combustion carbon capture plant. It showed that LSTM models are highly effective in forecasting amine emissions and critical system performance parameters during plant operations. The forecasting models achieved exceptional accuracy up to 99.0% on the test set over an operational period and successfully captured both overall trends and abrupt fluctuations in the data. The forecasting models were validated using real operational data from the TCM, focusing on amine emissions of AMP and Piperazine measured via FTIR and IMR-MS techniques. In addition, key system performance parameters such as $CO_2$ product flow, absorber outlet temperature before the WW, depleted FG outlet temperature, and RFCC stripper bottom temperature were also estimated by demonstrating the ML model's applicability across multiple plant output variables.

Unlike conventional mechanistic approaches, our method directly learns the complex mapping between plant input features and emissions behavior from operational data. The resulting models not only enable accurate real-time forecasting, but also offer interpretability, allowing us to identify the most influential parameters for emission mitigation and system optimization, covering multiple operational scenarios and input perturbations. Through causal impact analysis, we demonstrated that targeted interventions on key operational parameters can lead to measurable reductions in emissions and improvements in performance. This methodology illustrates how ML can serve not just as a forecasting tool, but also as a guide for operational decision-making under both steady-state and dynamic conditions.

Looking ahead, the proposed approach has broader implications beyond carbon capture applications. Industrial processes such as plant start-ups, fuel transitions, or solvent degradation phases generate large volumes of high-resolution data, which are often underutilized. Our findings suggest that machine learning models, particularly those incorporating active learning strategies, can transform this data into actionable insights, which accelerates time to operability by supporting process safety, and reducing reliance on prolonged empirical testing. Furthermore, the ability of machine learning to uncover underlying patterns in complex systems where mechanistic understanding is limited underscores the need for standardized, machine-actionable data sharing within the chemical engineering community. With continued development, such tools could revolutionize process optimization, making machine learning a cornerstone of future process design and operational excellence in the chemical and energy sectors.

**CRediT Authorship Contribution Statement**

**Lokendra Poudel:** Conceptualization, Methodology, Model design and optimization, Data analysis, Writing – original draft, Writing – review & editing, Supervision.
**David Tincher:** Model design and optimization, Data analysis, Writing – review & editing.
**Duy-Nhat Phan:** Model design and optimization, Data analysis, Writing – review & editing.
**Rahul Bhoumik:** Conceptualization, Resources, Writing – review & editing, Supervision.

**Declaration of Competing Interest**

Authors declare that they have no competing financial interests or personal connections that might have influenced the results of this research.

## Acknowledgement

This study is based upon research supported by the U.S. Department of Energy (DOE) Small Business Innovation Research (SBIR) Program, Phase I, under Contract No.: DE-SC0025013. Any opinions, findings, conclusions, or recommendations expressed in this study are those of the authors and do not necessarily reflect the views of the DOE. The authors would also like to acknowledge the U.S. DOE National Energy Technology Laboratory (NETL) and Technology Centre Mongstad (TCM) for providing data used in this study, which was collected TCM at in Mongstad, Norway. This acknowledgement also recognizes the TCM Owners, including the Norwegian state and the industrial partners Equinor, Shell, and TotalEnergies.

## Data and Code Availability

The authors do not have permission to share the data used in this study. Additionally, the developed codes are proprietary and owned by the company.